\documentclass[]{style/style}
\microtypesetup{expansion=false}

\usepackage{amsmath}
\usepackage{amssymb}
\usepackage{mathtools}
\usepackage{amsthm}
\usepackage{xspace}
\usepackage{colortbl}
\usepackage{multirow}
\usepackage{enumitem}
\usepackage{wrapfig}
\usepackage{array}
\usepackage{tabularx}

\newcommand{\method}{WanPE}
\newcommand{\bench}{WanPEval}
\graphicspath{{figures/}}
\makeatletter
\newcommand{\authornl}[2][]{%
  \begingroup
    \let\protect\@unexpandable@protect
    \xdef\authorlist{\authorlist\protect\\ \authorformat[#1]{#2}}%
  \endgroup
}
\newcommand{\affiliationnl}[2][]{%
  \begingroup
    \let\protect\@unexpandable@protect
    \xdef\affiliationlist{\affiliationlist\protect\\ \affiliationformat[#1]{#2}}%
  \endgroup
}
\makeatother

\title{WanPE: Towards Cinematic Prompt Enhancement for Modern Text-to-Video Generation}

\author[1,2*]{Yubo~Zhu}
\author[2,3*]{Yawen~Shao}
\author[2,4*]{Ziyun~Dai}
\author[2,3*]{Zixun~Fang}
\author[2\dagger\ddagger]{Kai~Zhu}
\author[2]{Siyang~Sun}
\authornl[2]{Haolan~Xue}
\author[2]{Chuxin~Wang}
\author[2]{Tingyu~Weng}
\author[2]{Jingming~Luo}
\author[2]{Chen~Shi}
\author[2]{Lianghua~Huang}
\author[2]{Yufeng~Ai}
\authornl[2]{Yuzheng~Wang}
\author[2]{Wenyuan~Zhang}
\author[2]{Yu~Shang}
\author[2]{Yuxiang~Bao}
\author[2]{Zoubin~Bi}
\author[2]{Jie~Xiao}
\author[2]{Jinbo~Xing}
\authornl[2]{Jiaxing~Zhao}
\author[2]{Chongyang~Zhong}
\author[2]{Hengjian~Chen}
\author[2]{Chenwei~Xie}
\author[2]{Akide~Liu}
\author[5]{Zhehan~Kan}
\authornl[2]{Yu~Liu}
\author[3\ddagger]{Wei~Zhai}
\author[1]{Sheng~Zhong}
\author[1\ddagger]{Wei~Tong}

\affiliation[1]{Nanjing University}
\affiliation[2]{Wan Team, Alibaba Group}
\affiliationnl[3]{University of Science and Technology of China}
\affiliation[4]{Fudan University}
\affiliation[5]{Tsinghua University}

\renewcommand\contribution[2][]{\addtolist[#1]{#2}{\contributionlist}{\contributionformat}{,\quad}}
\contribution[*]{Equal contribution}
\contribution[\dagger]{Project leader}
\contribution[\ddagger]{Corresponding author}

\abstract{
Video generation begins in text space by authoring a cinematic screenplay, then materializes into pixels. As contemporary video generators scale to $30$ seconds and faithfully follow complex conditions, the textual prompt largely directs the production, planning how actions, camera trajectories, lighting, and sound unfold across multi-shot sequences. In this paper, we present \textbf{\method{}}, a 397B-parameter prompt enhancement model trained on $1.05\mathrm{M}$ real-world videos to master director-level cinematic planning. \method{} formulates shot-level cinematic plans via video-grounded reverse construction, and employs Semantic-Consistency GRPO (SC-GRPO) to faithfully preserve user requirements across shots and over time. To benchmark this capability, we curate \textbf{\bench{}}, a human-annotated testbed covering durations from $5$ to $30$ seconds across varying intent granularities, supported by \(\sim\!11\)K blind pairwise assessments. When powering Wan3.0's video generator, \method{}-397B boosts human preference over raw user prompts by $10.66$-$18.84$ points at $5$-$15$ seconds, and by a dramatic $50.86$ points in the $30$-second arena. Ablation studies show that reverse construction demonstrates clear superiority over forward rewriting, while SC-GRPO robustly preserves semantic fidelity across model scales. Ultimately, \method{} leads all evaluated commercial offerings at $5$-$15$ seconds and remains competitive with Seedance~2.5 at $30$ seconds.

Project page: \url{https://wan-pe.github.io/}
}

\begin{document}

\maketitle
\makeatletter
{\renewcommand{\@makefntext}[1]{\noindent #1}\footnotetext{Further acknowledgements are detailed in the Acknowledgements section.}}
\makeatother

\section{Introduction}

Recent advances in text-to-video (T2V) generation have enabled systems such as Wan3.0~\citep{wanteam2026wanai} and Seedance 2.5~\citep{bytedance2026seedance25} to generate up to 30 seconds of cinematic-quality video, with expressive camera movements, realistic lighting, and coherent narratives. These systems rely on two tightly coupled components: a prompt enhancer that transforms user input into textual conditions, and a video generator that renders these conditions into video. As video generators process longer contexts and faithfully follow increasingly complex instructions, textual control now reaches the level of cinematic direction. Accordingly, the role of prompt enhancement shifts from enriching user requests with descriptive details to directing the entire production, planning how actions, camera, lighting, dialogue, and sound unfold coherently across shots and over time.

\begin{figure*}[htbp]
    \centering
    \includegraphics[width=0.95\textwidth]{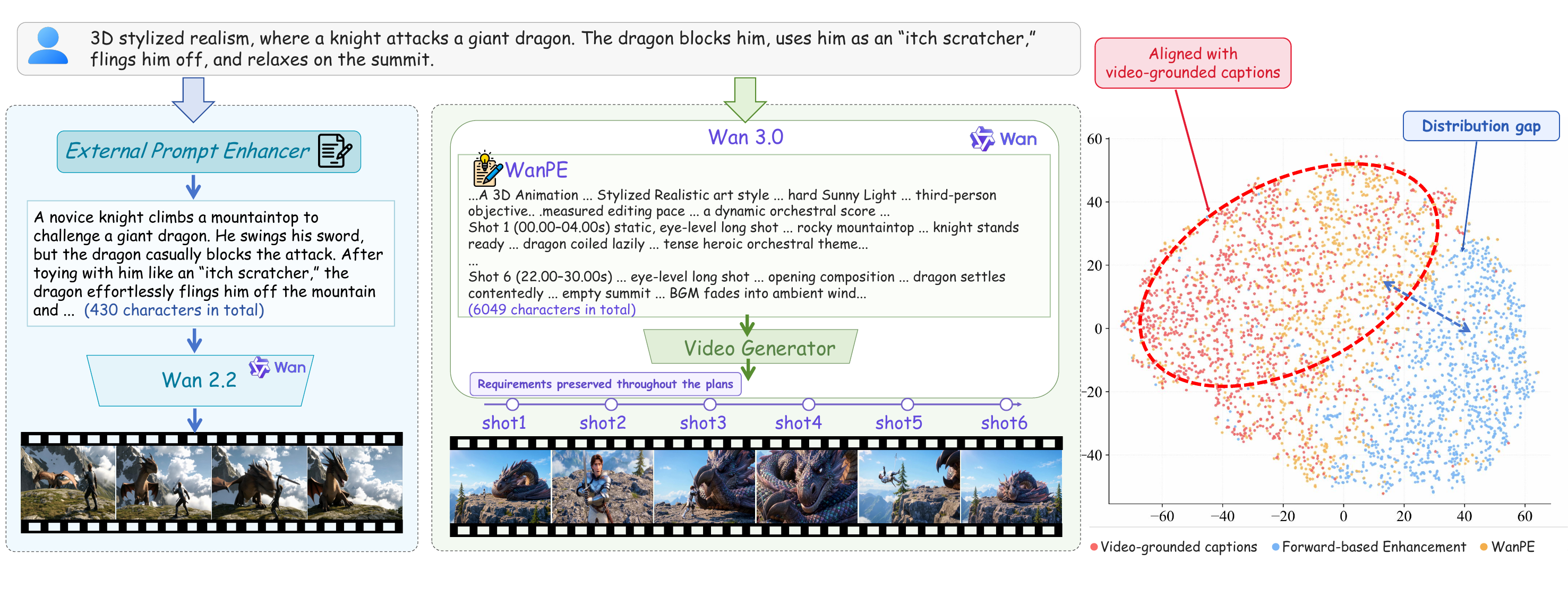}
    \vspace{-3mm}
    \caption{\textbf{Left:} Earlier models use optional prompt expansions, while modern video generators can process longer contexts and follow complex instructions. \textbf{Right:} t-SNE shows \method{} outputs align closely with video-grounded captions, while forward-based enhancement exhibits a distribution gap.}
    \vspace{-3mm}
    \label{fig:intro}
\end{figure*}

Many recent works have explored prompt enhancement for T2V generation, including learned prompt rewriting from user requests to detailed descriptions and multi-step LLM refinement~\citep{ijcai2025p1133,cheng2025vpo,ji2025prompt,wu2026phypromptrlbasedpromptrefinement,jia2026cape,Xue_2025_CVPR,ma2024pospromptsoptimizationsuite,yang-etal-2026-scmapr,gao2026rapocrossstagepromptoptimization,Long_2026_CVPR}. Historically, constrained by bounded context and weak instruction following in early video models, such as Wan2.2~\citep{wan2025wanopenadvancedlargescale}, prompt enhancement merely served as cosmetic visual enrichment. Today, modern video generators have broken these bottlenecks, stretching horizons to 30 seconds and unlocking unprecedented headroom for expressive textual control. In this paper, we break away from the conventional paradigm of descriptive prompt rewriting. Instead, we conceptualize text as the living flow of the video itself, the very blueprint where temporal cadence, camera choreography, and multi-shot transitions unfold before pixel rendering. Guided by this philosophy, we fundamentally redesign the prompt enhancement framework into a cinematic planning architecture. Realizing this architecture introduces two key challenges: learning coherent cinematic planning across interdependent production components, and preserving user-specified requirements throughout the unfolding plan.

For the first challenge, most T2V prompt enhancers are built around forward expansion from user requests to richer conditions. Their outputs follow a synthetic rewriting distribution. This creates a training–inference mismatch for downstream video generators trained on video-grounded captions. As prompt enhancement scales toward coherent cinematic planning, this conditioning mismatch reflects an asymmetry beyond linguistic style, as illustrated in Figure~\ref{fig:intro}. For example, to expand a request for a two-person action scene, a forward enhancer must not only imagine a sequence of fast interactions, but also determine how camera movement, lighting, audio, and other elements evolve with these interactions over time. However, a professionally filmed action sequence embodies coordination across acting, choreography, cinematography, editing, lighting, and sound. These real-world videos provide realized cinematic structure and video-grounded distribution targets.

For the second challenge, coherent cinematic planning elevates semantic preservation into a global consistency requirement over the entire plan. The enhancer must faithfully propagate user-specified subjects, actions, dialogue, camera instructions, and event order to every relevant shot while preserving their bindings and temporal relationships across the full sequence. At the same time, newly introduced details must remain compatible with both the original request and preceding planning decisions. As the plan unfolds, requirements may otherwise be omitted, altered, assigned to the wrong subject or shot, or contradicted later.

In this paper, we introduce \textbf{\method{}}, a prompt enhancer trained on \(1.05\mathrm{M}\) real-world videos for director-level cinematic planning. \method{} reverses conventional supervision construction by deriving a hierarchical cinematic condition \(y\) from each high-quality video and reconstructing a compatible user request \(x\). To preserve user requirements throughout the cinematic plan, we optimize the enhancer with SC-GRPO using a nine-dimensional reward that penalizes omissions, alterations, incorrect bindings, and temporal inconsistencies. We evaluate 4B--397B variants on \textbf{\bench{}}, a human-annotated 5-30-second testbed with varying intent granularities, using text-level scoring and \(\sim\!11\)K blind pairwise video assessments. \method{}-397B improves over raw prompts by up to $18.84$ points at 5-15 seconds and $50.86$ points at $30$ seconds. It leads evaluated commercial offerings at 5-15 seconds while remaining competitive with Seedance~2.5 at $30$ seconds. Ablations show reverse construction outperforms forward rewriting, while SC-GRPO improves semantic consistency by $18.6$-$23.3$ points across scales. \method{} transfers across video generators after format adaptation.

Our contributions are summarized as follows:
\vspace{-2mm}
\begin{itemize}
\item We redesign prompt enhancement as a scalable cinematic planning architecture, transforming text into the blueprint that orchestrates actions, camera choreography, audio, and narrative across shots and through time before pixel rendering.
\item We realize this architecture by learning realized cinematic structure from real-world videos through video-grounded reverse SFT and faithfully preserving all user requirements throughout the unfolding cinematic plan through SC-GRPO.
\item We introduce WanPEval, spanning diverse request complexities and generation durations from 5 to 30 seconds, and scale WanPE from 4B to 397B to demonstrate consistent effectiveness across model scales and downstream video generators.
\end{itemize}

\section{Methodology}

In this section, we present how WanPE realizes director-level cinematic planning while preserving user requirements across shots and over time. WanPE first formalizes these dual objectives (Section~\ref{sec:problem}), learns coherent cinematic planning through video-grounded reverse construction (Section~\ref{sec:video_grounded_sft}), and strengthens semantic fidelity with SC-GRPO (Section~\ref{sec:semantic_grpo}). Finally, WanPEval enables text and video-level evaluation across request granularities and generation durations (Section~\ref{sec:evaluation}).

\subsection{Problem Overview and Formulation}
\label{sec:problem}

A modern text-to-video (T2V) system composes a prompt enhancer
\(\pi_\theta\) with a video generator \(G_\phi\).
Given a natural-language user request \(x\) from the user-prompt distribution \(p_{\mathrm{u}}\), the prompt enhancer \(\pi_\theta\) produces
a textual condition \(y\), which guides the video generator \(G_\phi\) to synthesize
a video \(\hat{v}\):
\begin{equation}
x\sim p_{\mathrm{u}},\qquad
y\sim\pi_\theta\left(\cdot\mid x\right),\qquad
\hat{v}\sim G_\phi\left(\cdot\mid y\right).
\label{eq:t2v_pipeline}
\end{equation}

\textbf{The textual condition \(y\) should preserve all requirements in the user request \(x\).} For a given \(x\), multiple outputs may satisfy these requirements. We denote the set of such outputs by 
\(\mathcal{Y}_{\mathrm{sem}}\left(x\right)\):
\begin{equation}
\mathcal{Y}_{\mathrm{sem}}\left(x\right)
=
\left\{
y \;\middle|\;
c\left(y\right)=1,\ \forall c\in\mathcal{C}\left(x\right)
\right\},
\label{eq:semantic_feasible_set}
\end{equation}
where \(\mathcal{C}\left(x\right)\) denotes the set of user-specified semantic
and instructional constraints in \(x\), and \(c\left(y\right)=1\) indicates that \(y\) satisfies
the requirement \(c\) without omission, alteration, or
contradiction.

\textbf{The output distribution of the enhancer should align with the conditioning distribution of the video generator \(G_\phi\).} The video generator \(G_\phi\) is trained on real-world videos paired with video-grounded captions. We denote the video-grounded caption distribution by $p_{\mathrm{vg}}\left(y\right)$.

Combining semantic preservation with distributional alignment, we define
the ideal target distribution as the video-grounded caption distribution
restricted to semantically valid outputs:
\begin{equation}
p^{*}\left(y\mid x\right)
=
\frac{
p_{\mathrm{vg}}\left(y\right)
\mathbb{I}\!\left[y\in\mathcal{Y}_{\mathrm{sem}}\left(x\right)\right]
}{
Z\left(x\right)
},
\label{eq:target_distribution}
\end{equation}
where $Z\left(x\right)$ normalizes the distribution. We then learn \(\pi_\theta\) to match \(p^*\) over \(x\sim p_{\mathrm{u}}\):
\begin{equation}
\theta^{*}
=
\operatorname*{arg\,min}_{\theta}
\mathbb{E}_{x\sim p_{\mathrm{u}}}
\left[
D_{\mathrm{KL}}
\left(
p^{*}\left(\cdot\mid x\right)
\,\|\,\pi_\theta\left(\cdot\mid x\right)
\right)
\right].
\label{eq:pe_objective}
\end{equation}

\begin{figure*}[htbp]
    \centering
    \includegraphics[width=\textwidth]{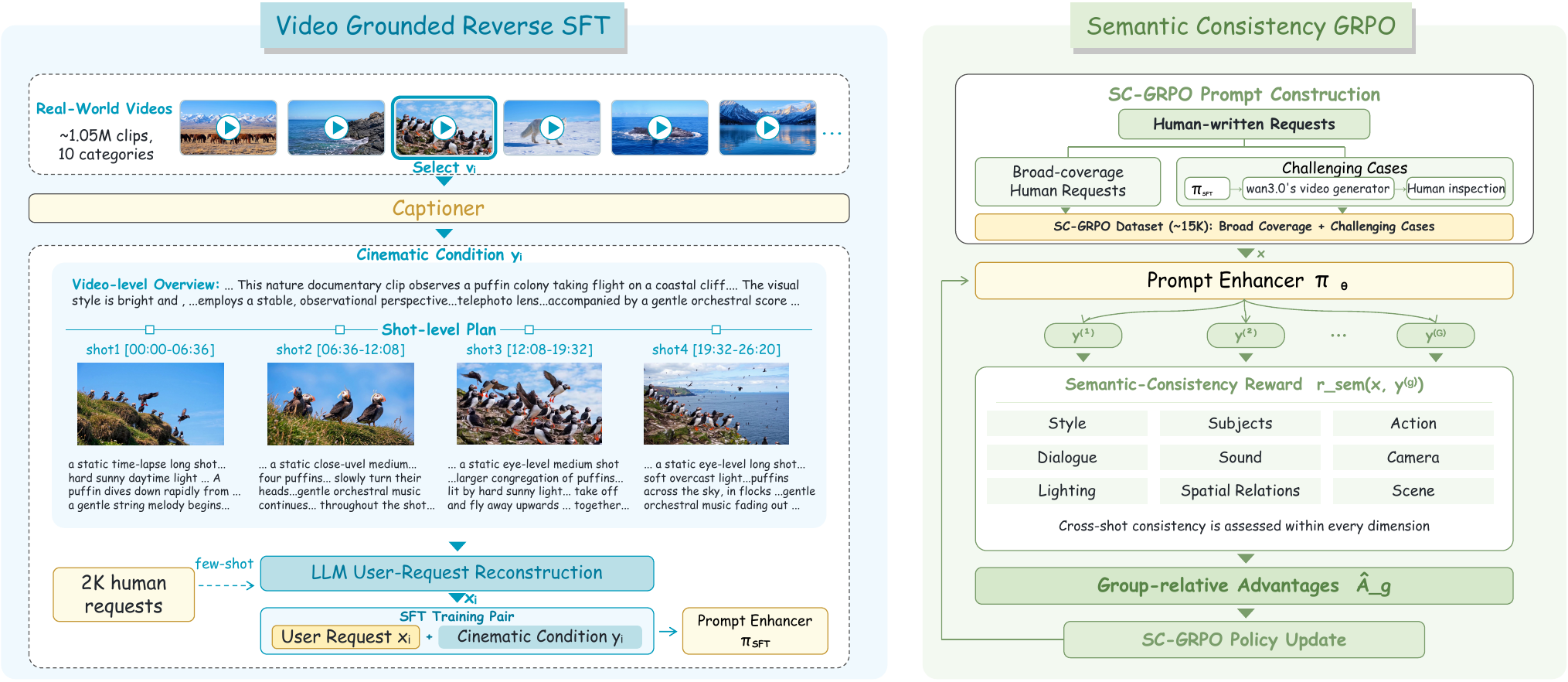}
    \vspace{-5mm}
    \caption{WanPE training pipeline. Video-grounded reverse SFT and semantic-consistency GRPO.}
    \vspace{-3mm}
\end{figure*}

Modern video generators can synthesize videos of up to 30 seconds, process longer textual contexts, and follow more complex instructions. This conditioning capacity creates unprecedented room for textual control. To harness this capacity, we redesign the prompt enhancement framework into a cinematic planning architecture. Learning such conditions poses two challenges. \textbf{(i) Learning coherent cinematic planning.} \(\pi_\theta\) must learn the joint organization of interdependent cinematic elements while matching \(p_{\mathrm{vg}}\). \textbf{(ii) Maintaining long-range semantic fidelity.} User-specified constraints such as identity, dialogue, and camera must stay consistent across temporally ordered shots.

\subsection{Video-Grounded Supervised Fine-Tuning}
\label{sec:video_grounded_sft}

To learn how cinematic elements are jointly organized across shots and over time while aligning the enhancer's outputs with \(p_{\mathrm{vg}}\), we first perform supervised fine-tuning on prompt--caption pairs \((x,y)\) satisfying \(y\in\mathcal{Y}_{\mathrm{sem}}(x)\). Starting from \(x\) and enriching it along predefined descriptive dimensions can readily satisfy this semantic requirement. However, the added content and its cinematic organization are determined by the rewriting procedure, causing the resulting targets to follow a synthetic rewriting distribution rather than \(p_{\mathrm{vg}}\). Our key design is to reverse the pair-construction process: we first obtain \(y\) by captioning a real-world video and then derive a semantically compatible \(x\) from \(y\).

\noindent\textbf{Real-world video collection and filtering.} To support alignment with \(p_{\mathrm{vg}}\), we collect a large and diverse set of real-world videos that are either publicly available or licensed for use. We segment long-form videos into clips of at most \(30\) seconds and apply multi-stage filters for technical validity,
visual quality, and motion quality. After filtering, the resulting dataset contains \(\sim\!1.05\mathrm{M}\) clips spanning ten content
dimensions with diverse camera movements. The details are in the Appendix~\ref{app:data_collection}.

\noindent\textbf{Video-grounded cinematic captioning.}
For each curated video clip \(v_i\), we use a multimodal video captioner \(f_{\mathrm{cap}}\) to analyze its visual and audio content and produce a textual cinematic target \(y_i\):
\begin{equation}
y_i=f_{\mathrm{cap}}(v_i;s_i),
\label{eq:video_grounded_captioning}
\end{equation}
where \(s_i\) denotes the category-specific captioning instruction for \(v_i\).

The target \(y_i\) organizes \(v_i\) hierarchically into a video-level summary and temporally ordered shot-level descriptions with timestamps. Each shot specifies its composition, subjects, actions, lighting, camera movement, transitions, dialogue, music, and sound effects, together with their progression over time. The instruction \(s_i\) adapts this structure to the category of \(v_i\). We keep captions that pass checks for structural completeness, timestamp validity, and consistency with the source video, and use them as video-grounded targets from
\(p_{\mathrm{vg}}\) for SFT. The details are in the Appendix~\ref{app:reverse_construction}.

\noindent\textbf{User-request reconstruction.}
In reverse pair construction, we derive a user request \(x_i\) from
video-grounded target \(y_i\). Summarizing \(y_i\) tends to retain the
caption's structure and detail, producing a compressed caption rather than
a natural user request. We therefore employ
\texttt{gpt-5.4} with few-shot prompting, using a pool of
\(2\mathrm{K}\) requests written by human annotators across ten content
categories.

Let \(d_i\) denote the category of \(y_i\), and let
\(\mathcal{H}_{d_i}\) denote the corresponding request pool. We sample five
demonstrations \(\mathcal{E}_i\) from \(\mathcal{H}_{d_i}\). Given \(y_i\)
and \(\mathcal{E}_i\), the LLM reconstructs
\begin{equation}
x_i=f_{\mathrm{LM}}(y_i;\mathcal{E}_i).
\label{eq:user_request_reconstruction}
\end{equation}

The reconstruction prompt requires \(x_i\) to contain only requirements
supported by \(y_i\), ensuring \(y_i\in\mathcal{Y}_{\mathrm{sem}}(x_i)\).
The demonstrations \(\mathcal{E}_i\) guide \(x_i\) toward language and
specificity of natural user requests.

\noindent\textbf{Supervised fine-tuning.}
The reverse construction yields the SFT dataset
\(\mathcal{D}_{\mathrm{SFT}}=\{(x_i,y_i)\}_{i=1}^{N}\). We fine-tune the
prompt enhancer \(\pi_\theta\) by minimizing the negative log-likelihood
of each video-grounded cinematic condition given its reconstructed user
request:
\begin{equation}
\mathcal{L}_{\mathrm{SFT}}(\theta)
=
-\frac{1}{N}
\sum_{i=1}^{N}
\log\pi_\theta(y_i\mid x_i).
\label{eq:sft_objective}
\end{equation}

Through this objective, the enhancer learns to transform natural user
requests into video-grounded cinematic conditions that capture the joint
organization of elements across shots and over time.

\subsection{Semantic-Consistency GRPO}
\label{sec:semantic_grpo}

To preserve user requirements as cinematic conditions unfold across shots and events, we introduce Semantic-Consistency GRPO (SC-GRPO). It applies Group Relative Policy Optimization~\citep{shao2024deepseekmathpushinglimitsmathematical} with a semantic-consistency reward penalizing omissions, alterations, inconsistent bindings between subjects, actions, and dialogue, and temporal inconsistencies across shots.

\noindent\textbf{Training data construction.}
To cover diverse requests and challenging cases, human annotators write T2V prompts for videos up to 30 seconds. We generate videos with Wan3.0's video generator conditioned on the SFT enhancer's outputs. Through visual assessment, annotators flag prompts whose videos inadequately realize the requested content. We combine these challenging cases with diverse human-written prompts to form \(\mathcal{D}_{\mathrm{SC\text{-}GRPO}}\), containing approximately \(15\mathrm{K}\) prompts.

\noindent\textbf{Semantic consistency reward.} We use \texttt{Qwen3.7-Max} to assess whether \(y\) preserves the specified constraints \(\mathcal{C}(x)\), producing a text-only reward \(r_{\mathrm{sem}}(x,y)\in[0,100]\), with higher scores indicating better requirement preservation. Evaluation spans nine dimensions: style, subjects, actions, dialogue, sound, camera, lighting, spatial relations, and scene. The evaluator checks for omitted, weakened, altered, or contradictory user requirements, incorrect subject--attribute and speaker--dialogue bindings, action and shot ordering errors, and cross-shot conflicts. Semantically equivalent paraphrases and compatible elaborations are accepted, while semantic discrepancies are penalized by severity.

\noindent\textbf{Policy optimization.}
We initialize \(\pi_\theta\) from the frozen reference \(\pi_{\mathrm{SFT}}\). For each \(x\in\mathcal{D}_{\mathrm{SC\text{-}GRPO}}\), we sample \(G\) conditions \(\{y^{(g)}\}_{g=1}^{G}\) from \(\pi_{\theta_{\mathrm{old}}}(\cdot\mid x)\) and standardize their rewards \(r_{\mathrm{sem}}(x,y^{(g)})\) into group-relative advantages \(\hat A_g\). We maximize
\begin{equation}
\mathcal J_{\mathrm{SC\text{-}GRPO}}(\theta)
=\mathbb E\!\left[
\frac{1}{G}\sum_{g=1}^{G}\frac{1}{T_g}\sum_{t=1}^{T_g}
\left(
\min\!\left\{\rho_{g,t}\hat A_g,
\operatorname{clip}(\rho_{g,t},1-\epsilon,1+\epsilon)\hat A_g\right\}
-\beta\mathcal K_{g,t}
\right)
\right],
\label{eq:sc_grpo_objective}
\end{equation}
where \(T_g\) is output length, \(\rho_{g,t}\) the current/old policy ratio, and \(\mathcal K_{g,t}\) the SFT-reference KL penalty.

\subsection{WanPEval: Testbed and Evaluation}
\label{sec:evaluation}

Evaluating modern T2V prompt enhancers requires diverse generation durations and request complexities. We therefore introduce \textbf{\bench{}}, a human-annotated testbed spanning \(5\) to \(30\) seconds and requests ranging from concise high-level intents to detailed shot-level instructions.

\noindent\textbf{Data construction and curation.}
Human annotators write practical T2V requests across diverse topics, stratified by duration and granularity, from high-level intents to structured shot-level instructions. We review requests for semantic clarity, internal consistency, and temporal feasibility, remove duplicates, and exclude overlap with SFT and SC-GRPO training data. The curated \bench{} contains 249 requests. Details are in the Appendix~\ref{app:benchmark_design}.

\noindent\textbf{Text-level semantic evaluation.} To evaluate semantic fidelity
at text level, we apply the semantic consistency reward
\(r_{\mathrm{sem}}(x,y)\) defined in Section~\ref{sec:semantic_grpo} to each
condition generated for \bench{}.

\noindent\textbf{Expert video evaluation.}
The practical value of a prompt enhancer lies in the quality of downstream videos conditioned on its outputs, which requires jointly assessing request adherence, perceptual quality, and cinematic coherence. As automated metrics struggle to reliably capture these aspects, we adopt anonymous pairwise human preference evaluation following the Artificial Analysis Video Arena~\citep{artificialanalysis_video}, a widely recognized leaderboard for modern video generation models.

Each of the \(M\) methods generates one video per request, yielding up to \(N\binom{M}{2}\) pairs across \(N\) requests. Evaluation involves \(60\) experts in screenwriting, directing, cinematography, and related film disciplines. Each pair is presented with its request \(x\), with method identities hidden and left--right order randomized. Experts select one of four outcomes: A preferred, B preferred, both good, or both poor. For method \(k\), let \(N_k\) denote its number of valid comparisons after excluding unsuccessful generations, and let \(N_{\mathrm{win}}^k\) and \(N_{\mathrm{bg}}^k\) denote its win and both-good counts, respectively. We compute its lower-bound, upper-bound, and overall preference scores as
\[
S_{\mathrm{lower}}^k=\frac{N_{\mathrm{win}}^k}{N_k}, \qquad
S_{\mathrm{upper}}^k=\frac{N_{\mathrm{win}}^k+N_{\mathrm{bg}}^k}{N_k}, \qquad
S^k=\frac{N_{\mathrm{win}}^k+0.5N_{\mathrm{bg}}^k}{N_k}.
\]

We additionally fit a Bradley--Terry model~\citep{bradley1952rank} to account for opponent strength, treating both-good and both-poor outcomes as ties:
\[
p_{kj}=\sigma(b_k-b_j),\qquad
\sum_k b_k=0,\qquad
\mathrm{BT}^k=100\,\sigma(b_k).
\]

Here, \(b_k\) denotes strength. We report \(\mathrm{BT}^k\) as the preference score against a mean-strength opponent.

\section{Experiments}

In this section, we comprehensively evaluate \method{} from four perspectives: generation quality (Section~\ref{sec:overall_comparison}), the effectiveness of reverse-constructed supervision (Section~\ref{sec:forward_api_comparison}), transferability across downstream video generators (Section~\ref{sec:cross_generator_generalization}), and training analysis (Section~\ref{sec:semantic_analysis}).

\textbf{Settings.} We instantiate four model variants initialized from Qwen3.5-4B, Qwen3.5-9B, Qwen3.5-35B-A3B, and Qwen3.5-397B-A17B~\citep{qwen35}, denoted as \method{}-4B, \method{}-9B, \method{}-35B, and \method{}-397B, respectively. The experiments are conducted on 512 GPUs. Unless otherwise specified, each request is first enhanced by \method{}, and the output is then provided to Wan3.0's video generator~\citep{wanteam2026wanai} for video generation and evaluation. For an evaluation involving $n$ methods, we construct all $\binom{n}{2}$ pairwise video comparisons for each request and aggregate the resulting judgments to compute the final scores.

\subsection{Overall Comparison}
\label{sec:overall_comparison}

We evaluate the generation quality of \method{} on \bench{}. We compare our prompt-enhanced pipeline with Seedance 2.5~\citep{bytedance2026seedance25}, Seedance 2.0~\citep{seedance2026seedance20advancingvideo}, HappyHorse 1.1~\citep{happyhorse}, Kling 3.0~\citep{klingai2026}, MiniMax-H3~\citep{minimaxh3}, and LTX-2.5~\citep{ltx25}, and an original-request baseline without prompt enhancement. Given each
system's maximum supported duration, we evaluate Seedance 2.5 against \method{}-397B on the 30-second subset and all other systems on the 5--15-second subset of \bench{}.

\paragraph{Settings.} For MiniMax-H3 and LTX-2.5, each request is first enhanced by the native prompt enhancer, H3-Context-IR or LTX-2.5-PE, and the enhanced condition is passed to the generator, H3-Base or LTX-2.5-Base. For the remaining baselines, the native prompt enhancer and video generator form an end-to-end API, so we submit each request directly and evaluate the returned video.

\begin{table}[t]
\centering
\scriptsize
\setlength{\tabcolsep}{1.1pt}
\renewcommand{\arraystretch}{1.12}
\caption{Expert preference score and Bradley--Terry score on the $5$--$15$ seconds subset of \bench{}.}
\begin{tabular*}{\columnwidth}{
    @{\extracolsep{\fill}}l|cc cc cc cc@{}
}
\toprule
& \multicolumn{2}{c}{5 seconds}
& \multicolumn{2}{c}{10 seconds}
& \multicolumn{2}{c}{15 seconds}
& \multicolumn{2}{c}{Overall} \\
\cmidrule(lr){2-3}
\cmidrule(lr){4-5}
\cmidrule(lr){6-7}
\cmidrule(l){8-9}
Method
& \(S\) & BT
& \(S\) & BT
& \(S\) & BT
& \(S\) & BT \\
\midrule

\shortstack[l]{LTX-2.5}
& 18.69 & 31.23
& 18.47 & 34.74
& 16.10 & 32.20
& 17.30 & 33.02 \\

Kling 3.0
& 31.11 & 43.02
& 23.47 & 39.84
& 16.10 & 34.07
& 20.80 & 37.40 \\

HappyHorse 1.1
& 20.28 & 32.82
& 24.00 & 40.72
& 26.80 & 42.39
& 24.89 & 40.46 \\

\shortstack[l]{MiniMax-H3}
& 29.09 & 41.24
& 36.46 & 49.59
& 38.30 & 51.13
& 36.40 & 49.27 \\

Seedance 2.0
& 46.20 & 54.20
& 41.40 & 52.16
& 44.31 & 56.90
& 43.42 & 54.70 \\

\rowcolor{gray!12}
\multicolumn{9}{c}{\textit{downstream generator: Wan3.0's video generator}} \\

Original request
& 46.08 & 57.13
& 33.63 & 49.40
& 30.59 & 47.69
& 33.80 & 49.59 \\

+\method{}-4B
& 48.17 & 57.28
& 41.27 & 54.58
& 44.00 & 57.10
& 43.60 & 56.21 \\

+\method{}-9B
& \underline{50.91} & \underline{59.76}
& 44.40 & 57.06
& 45.78 & 58.24
& 46.00 & 58.01 \\

+\method{}-35B
& 50.69 & 59.32
& \underline{48.76} & \underline{60.22}
& \underline{46.45} & \underline{60.27}
& \underline{47.85} & \underline{60.09} \\

+\method{}-397B
& \textbf{56.74} & \textbf{65.00}
& \textbf{49.91} & \textbf{62.00}
& \textbf{49.43} & \textbf{60.95}
& \textbf{50.61} & \textbf{61.85} \\
\bottomrule
\end{tabular*}
\label{tab:short_medium_result}
\end{table}

\begin{figure*}[htbp]
    \centering
    \vspace{-4mm}
    \includegraphics[width=\textwidth]{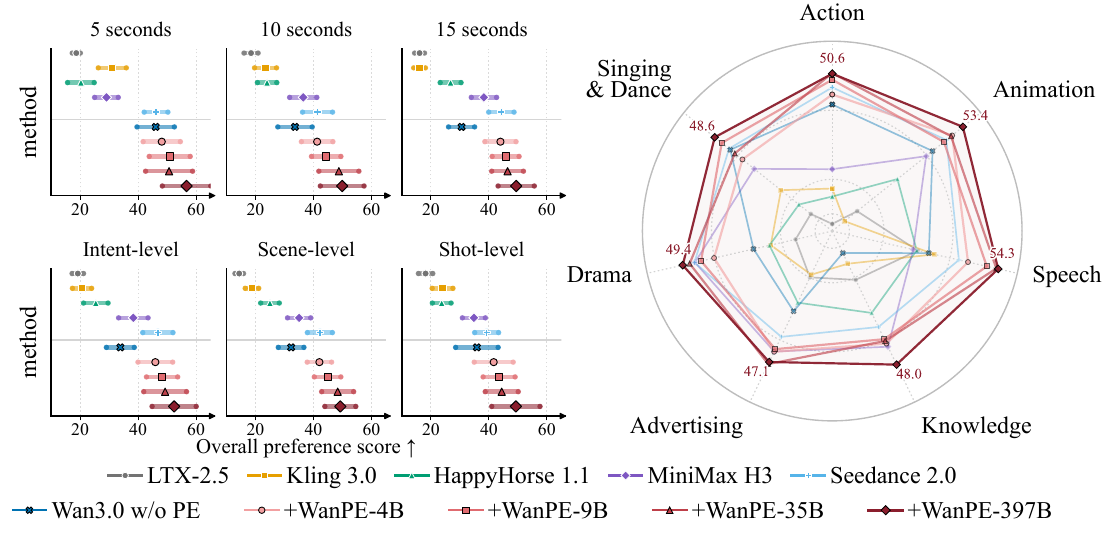}
    \vspace{-8mm}
    \caption{Fine-grained expert evaluation on the $5$-$15$ seconds subset of \bench{}. \textbf{Left:} Preference scores across generation durations and request-granularity levels. Markers denote \(S\), and horizontal intervals span \([S_{\mathrm{lower}}, S_{\mathrm{upper}}]\). \textbf{Right:} Preference scores \(S\) across seven content categories.}
    \label{fig:\bench{}_breakdown}
\end{figure*}

\begin{table}[t]
\centering
\scriptsize
\setlength{\tabcolsep}{1.6pt}
\renewcommand{\arraystretch}{1.08}
\caption{Preference scores \(S\) on \bench{}.
(i)~30-second subset against Seedance~2.5, under Wan3.0.
(ii)~Reverse-constructed vs.\ forward-based enhancement, under Wan3.0.
(iii)~Format-adapted \method{} vs.\ native enhancers on LTX-2.5 and MiniMax-H3 in two separate battles.}
\begin{tabular*}{\columnwidth}{@{\extracolsep{\fill}}l|cccccccc@{}}
\toprule
Method
& Action & Anim. & Speech & Ad.
& Sing \& Dance & Drama & Know. & Overall \\
\midrule
\rowcolor{blue!10}
\multicolumn{9}{c}{Exp1: 30-second subset} \\
Seedance 2.5
& \textbf{68.18}
& \underline{46.43}
& \underline{55.00}
& \textbf{81.25}
& \textbf{60.00}
& \textbf{56.00}
& \textbf{60.71}
& \underline{59.76} \\
\rowcolor{gray!12}
\multicolumn{9}{c}{\textit{downstream generator: Wan3.0's video generator}} \\
Original request
& 4.17
& 9.38
& 15.79
& 5.00
& 22.50
& 4.00
& 3.57
& 9.38 \\
+\method{}-397B
& \underline{50.00}
& \textbf{81.25}
& \textbf{73.68}
& \underline{75.00}
& \underline{47.50}
& \underline{53.85}
& \underline{57.14}
& \textbf{60.24} \\
\midrule
\rowcolor{blue!10}
\multicolumn{9}{c}{Exp2: Reverse vs.\ forward enhancement} \\
\rowcolor{gray!12}
\multicolumn{9}{c}{\textit{downstream generator: Wan3.0's video generator}} \\
Original Request
& 23.97 & 21.01 & 29.85 & 19.32
& 35.00 & 22.60 & 6.38 & 23.05 \\
+ Forward Rewriting
& \underline{39.57} & 38.43 & \underline{42.13} & \underline{43.57}
& \underline{38.61} & \underline{34.62} & \underline{41.89} & \underline{39.49} \\
+ Forward-target SFT
& 30.36 & \underline{44.23} & 32.28 & 37.00
& 35.34 & 33.72 & 34.26 & 35.17 \\
+ \method{}-397B-SFT
& \textbf{46.48} & \textbf{55.98} & \textbf{48.31} & \textbf{49.32}
& \textbf{45.51} & \textbf{51.54} & \textbf{51.23} & \textbf{49.86} \\
\midrule
\rowcolor{blue!10}
\multicolumn{9}{c}{Exp3: Cross-generator transfer (5--15s)} \\
\rowcolor{gray!12}
\multicolumn{9}{c}{\textit{downstream generator: LTX-2.5-Base}} \\
+ LTX-2.5-PE
& 8.33 & 18.33 & 30.00 & \textbf{32.50}
& 7.50 & 26.67 & 25.00 & 21.11 \\
+ \method{}-397B
& \textbf{25.00} & \textbf{38.33} & \textbf{53.33} & 27.50
& \textbf{27.50} & \textbf{36.67} & \textbf{35.00} & \textbf{35.56} \\
\rowcolor{gray!12}
\multicolumn{9}{c}{\textit{downstream generator: MiniMax-H3-Base}} \\
+ H3-Context-IR
& \textbf{44.44} & 31.03 & 33.33 & 39.47
& 37.50 & 29.31 & \textbf{40.00} & 35.92 \\
+ \method{}-397B
& 29.63 & \textbf{37.93} & \textbf{40.00} & \textbf{44.74}
& \textbf{47.50} & \textbf{50.00} & \textbf{40.00} & \textbf{41.09} \\
\bottomrule
\end{tabular*}
\label{tab:combined_s_ablations}
\vspace{-1mm}
\end{table}

\textbf{\method{} is strong and consistent on \bench{}.} We first ablate PE scale under the same Wan3.0 video generator. Figure~\ref{fig:\bench{}_breakdown} shows that \method{}-397B ranks first on intent-, scene-, and shot-level requests, with scores of \(52.44\), \(49.26\), and \(49.32\). Compared with using no prompt enhancement, \method{}-397B delivers substantial gains of \(18.55\), \(16.97\), and \(13.45\) points across the three request granularities. We then compare \method{}-397B with other leading video generation systems. As shown in Table~\ref{tab:short_medium_result}, it ranks first in both \(S\) and BT at every duration on the 5--15-second subset, with overall scores of \(50.61\) and \(61.85\), and outperforms Seedance 2.0 by \(7.19\) points in \(S\). As shown in Table~\ref{tab:combined_s_ablations}, it remains competitive with Seedance 2.5 on the 30-second subset, especially on animation and speech, where it scores \(81.25\) and \(73.68\). Details are provided in the Appendix~\ref{app:additional_quantitative_results}.

\textbf{Prompt enhancement becomes increasingly important as the generation horizon grows.} Under the same Wan3.0 video generator, \method{}-397B improves over the original-request baseline by \(10.66\), \(16.28\), and \(18.84\) points at 5, 10, and 15 seconds. On the 30-second subset, it raises \(S\) from \(9.38\) to \(60.24\). The gains therefore increase over 5--15 seconds and remain large at 30 seconds.

\subsection{Reverse-Constructed Supervision vs. Forward Prompt Expansion}
\label{sec:forward_api_comparison}

A central design choice of \method{} is to construct supervision in reverse from video-grounded captions. We isolate its effect through two forward-based baselines: (i) \textbf{Forward Rewriting.} $8$ experts in computer science and film directing designed an instruction covering semantic fidelity, cinematic structure, temporal progression, camera language, lighting, narrative development, and audiovisual
coherence. These dimensions match those of the video-grounded targets \(y\) in Section~\ref{sec:video_grounded_sft}. On a separate validation
set, gemini-3.1-pro-preview~\citep{gemini31pro} rewrites the original requests, and the Wan3.0 DiT generates videos conditioned on the rewritten prompts. Following the blind protocol in
Section~\ref{sec:evaluation}, we refined the instruction over 27
versions. The final version is the Forward Rewriting baseline.
(ii) \textbf{Forward-target SFT.}
Using the reconstructed requests \(x\) from
Section~\ref{sec:video_grounded_sft}, we apply the above Forward Rewriting
method to obtain \(y_{\mathrm{fwd}}\), and train on \((x,y_{\mathrm{fwd}})\) with same settings.

\textbf{Video-grounded learning provides stronger conditions than forward rewriting.} As shown in Table~\ref{tab:combined_s_ablations}, \method{}-397B-SFT attains an overall \(S\) of \(49.86\), outperforming Forward Rewriting by \(10.37\) points and Forward-target SFT by \(14.69\) points. It ranks first in all seven content categories. These results indicate that video-grounded cinematic captions provide more effective conditions than forward-constructed targets.

\subsection{Transferability of Cinematic Planning Across Video Generators}
\label{sec:cross_generator_generalization}

We test whether \method{} plans remain effective beyond Wan3.0. Because video generators use different conditioning formats, we adapt \method{} to each native format while keeping the cinematic content. We use MiniMax-H3 and LTX-2.5, whose enhancers and generators are separately accessible. GPT-5.4 converts \method{} outputs into H3-Context-IR and LTX-2.5-PE formats, then fed to H3-Base and LTX-2.5-Base.

\textbf{\method{} transfers across video generators.}
After format adaptation, \method{}-397B outperforms the native enhancers on both generators (Table~\ref{tab:combined_s_ablations}, Exp3), improving \(S\) by \(14.45\) points on LTX-2.5-Base and \(5.17\) on MiniMax-H3-Base. These results suggest its learned cinematic organization remains effective across formats.

\begin{table}[t]
\centering
\scriptsize
\setlength{\tabcolsep}{1.8pt}
\renewcommand{\arraystretch}{1.08}
\newcommand{\gain}[1]{%
  \rlap{\hspace{1pt}{%
    \color{blue!75!black}\fontsize{4.6}{5.0}\selectfont $#1$%
  }}%
}
\caption{(i)~Semantic consistency evaluated by gemini-3.1-pro-preview. \(\dagger\) denotes results computed on the 5--15-second subset. (ii)~Expert preference \(S\) for 397B SFT vs.\ \method{}-397B under Wan3.0.}
\begin{tabular*}{\columnwidth}{@{\extracolsep{\fill}}lccccccc@{\hspace{16pt}}}
\toprule
\rowcolor{blue!10}
\multicolumn{8}{c}{\textit{(i) Semantic consistency}} \\
\midrule
Method
& Overall$\uparrow$
& 5 seconds$\uparrow$
& 10 seconds$\uparrow$
& 15 seconds$\uparrow$
& 30 seconds$\uparrow$
& Perfect$\uparrow$
& Failure$\downarrow$ \\
\midrule
Fwd. Rewriting
& 90.7 & 93.5 & 91.8 & 90.3 & 89.4 & 55.0 & 10.0 \\
LTX-2.5-PE
& 77.3 & 76.1 & 81.6 & 74.9 & 77.0 & 33.7 & 38.6 \\
H3-Context-IR
& $88.3^{\dagger}$ & 89.4 & 88.8 & 87.6 & --
& $54.2^{\dagger}$ & $17.3^{\dagger}$ \\
\midrule
\rowcolor{gray!12}
\multicolumn{8}{c}{\textit{Ours}} \\
\method{}-4B-SFT
& 66.5 & 67.0 & 68.4 & 66.7 & 64.2 & 20.1 & 56.6 \\
\quad + SC-GRPO
& 85.3\gain{+18.8}
& 91.9\gain{+24.9}
& 83.1\gain{+14.7}
& 85.3\gain{+18.6}
& 84.8\gain{+20.6}
& 52.0\gain{+31.9}
& 27.4\gain{-29.2} \\
\cmidrule(lr){1-8}
\method{}-9B-SFT
& 70.2 & 74.6 & 69.5 & 72.9 & 65.6 & 22.5 & 48.6 \\
\quad + SC-GRPO
& 88.8\gain{+18.6}
& 94.0\gain{+19.4}
& 90.6\gain{+21.1}
& 87.9\gain{+15.0}
& 86.5\gain{+20.9}
& 53.8\gain{+31.3}
& 15.7\gain{-32.9} \\
\cmidrule(lr){1-8}
\method{}-35B-SFT
& 73.1 & 69.0 & 70.1 & 77.3 & 71.7 & 21.3 & 42.6 \\
\quad + SC-GRPO
& \underline{96.4}\gain{+23.3}
& \textbf{99.3}\gain{+30.3}
& \underline{97.9}\gain{+27.8}
& \underline{95.0}\gain{+17.7}
& \underline{95.9}\gain{+24.2}
& \underline{80.7}\gain{+59.4}
& \underline{4.4}\gain{-38.2} \\
\cmidrule(lr){1-8}
\method{}-397B-SFT
& 75.5 & 73.2 & 70.0 & 81.5 & 73.4 & 29.7 & 36.9 \\
\quad + SC-GRPO
& \textbf{97.6}\gain{+22.1}
& \underline{99.1}\gain{+25.9}
& \textbf{98.0}\gain{+28.0}
& \textbf{96.9}\gain{+15.4}
& \textbf{97.6}\gain{+24.2}
& \textbf{85.5}\gain{+55.8}
& \textbf{2.8}\gain{-34.1} \\
\bottomrule
\end{tabular*}

\begin{tabular*}{\columnwidth}{@{\extracolsep{\fill}}l|cccccccc@{}}
\rowcolor{blue!10}
\multicolumn{9}{c}{\textit{(ii) PE-scale expert preference on \bench{}}} \\
\midrule
Method
& Action & Anim. & Speech & Ad.
& Sing \& Dance & Drama & Know. & Overall \\
\midrule
\rowcolor{gray!12}
\multicolumn{9}{c}{\textit{Video generator: Wan3.0's video generator}} \\
Original Request
& 24.71 & 26.62 & 27.44 & 11.00
& 32.50 & 30.12 & 15.74 & 24.95 \\
+ 397B SFT
& \underline{47.02} & \underline{47.37} & \underline{40.24} & \underline{54.00}
& \underline{37.50} & \underline{38.75} & \underline{34.26} & \underline{42.70} \\
+ \method{}-397B
& \textbf{47.65} & \textbf{48.70} & \textbf{53.05} & \textbf{55.00}
& \textbf{53.33} & \textbf{40.96} & \textbf{53.70} & \textbf{49.69} \\
\bottomrule
\end{tabular*}
\label{tab:consistency_and_scale}
\end{table}

\subsection{Semantic Consistency and SC-GRPO Analysis}
\label{sec:semantic_analysis}

\begin{wrapfigure}[11]{r}{0.3\columnwidth}
    \vspace{-10pt}
    \centering
    \includegraphics[width=\linewidth]{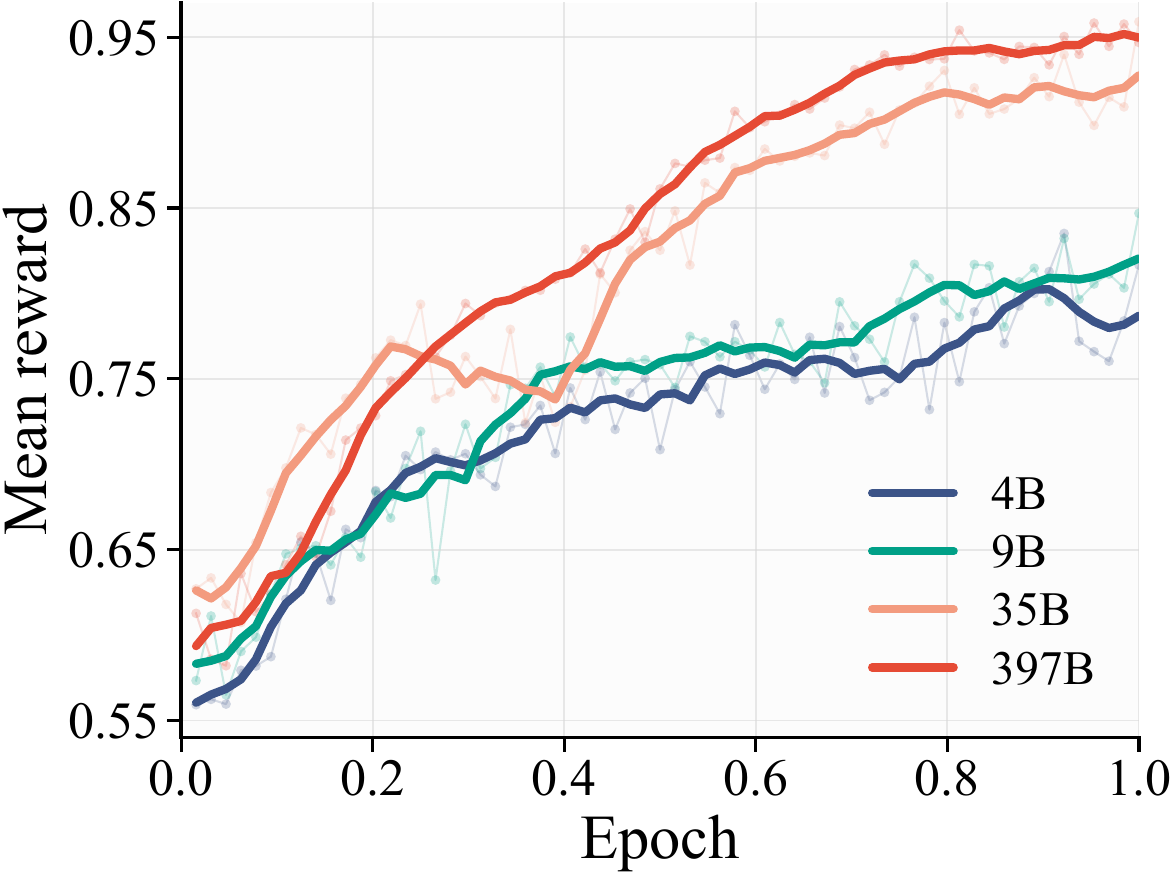}
    \vspace{-8mm}
    \caption{Mean reward during SC-GRPO training at four scales.}
    \label{fig:sc_grpo_training}
\end{wrapfigure}

In this subsection, we evaluate whether SC-GRPO improves semantic consistency and whether this improvement also benefits downstream video generation.

\textbf{SC-GRPO consistently improves semantic consistency.} Figure~\ref{fig:sc_grpo_training} shows that the mean training reward increases substantially across four model sizes, with larger models reaching higher final values. These gains are reflected on \bench{}. As shown in Table~\ref{tab:consistency_and_scale}, SC-GRPO improves the overall semantic-consistency score by \(18.6\)--\(23.3\) points across four model sizes, as evaluated by \texttt{gemini-3.1-pro-preview}, with consistent gains at every evaluated duration. It increases the proportion of perfect outputs and reduces failures.

\textbf{SC-GRPO improves downstream video generation.}
We compare \method{}-397B before and after SC-GRPO using the expert video-evaluation protocol in Section~\ref{sec:evaluation}. Both variants use the same Wan3.0 video generator and differ only in their enhanced conditions. As shown in Table~\ref{tab:consistency_and_scale}, SC-GRPO increases the overall preference score from 42.70 to 49.69 and improves performance across all seven content categories, with larger gains in knowledge, singing and dance, and speech. These results thus show that optimizing semantic consistency also benefits downstream video quality, producing more effective textual conditions for video generation.

\section{Related Work}

\paragraph{Text-to-video generation.}
Text-to-video generation has progressed rapidly with diffusion and transformer architectures, from early latent video models~\citep{blattmann2023stablevideodiffusionscaling,wang2023laviehighqualityvideogeneration,guo2024animatediffanimatepersonalizedtexttoimage,ma2025lattelatentdiffusiontransformer} to large diffusion transformers such as Open-Sora~\citep{zheng2024opensorademocratizingefficientvideo}, CogVideoX~\citep{yang2025cogvideoxtexttovideodiffusionmodels}, HunyuanVideo~1.5~\citep{wu2025hunyuanvideo15technicalreport}, Wan2.2~\citep{wan2025wanopenadvancedlargescale}, LTX series~\citep{hacohen2026ltx2efficientjointaudiovisual}, Lingbot-Video~\citep{ma2026scalingmixtureofexpertsvideopretraining}, and MiniMax-H3~\citep{minimaxh3}.
Closed-source systems include Sora~2~\citep{openai2025sora2}, Kling~\citep{klingai2026}, Seedance~2.5~\citep{bytedance2026seedance25}, Wan3.0~\citep{wanteam2026wanai}.

\paragraph{Prompt enhancement for text-to-video generation.}
Recent works have explored T2V prompt enhancement from several directions. POS~\citep{ma2024pospromptsoptimizationsuite} jointly optimizes noise and textual prompts with semantic-preserving rewriting, while RAPO~\citep{Gao_2025_CVPR} combines retrieval-augmented modifiers with LLM rewriting to better align user prompts with the training-prompt distribution. VPO~\citep{cheng2025vpo} and Prompt-A-Video~\citep{ji2025prompt} train prompt optimizers with supervised and preference-based objectives, while VidPrompter~\citep{ijcai2025p1133} adopts multimodal multi-task training with hallucination-aware preference optimization. CAPE-T2V~\citep{jia2026cape} studies captioner-anchored two-sided conditioning alignment. Other methods focus on iterative refinement during inference: PhyT2V~\citep{Xue_2025_CVPR} introduces reasoning-based self-refinement for physical realism, SCMAPR~\citep{yang-etal-2026-scmapr} coordinates scenario-aware agents with semantic verification, and VISTA~\citep{Long_2026_CVPR} iteratively revises prompts using generated-video selection and critique. RAPO++~\citep{gao2026rapocrossstagepromptoptimization} combines training-data alignment, test-time optimization, and rewriter fine-tuning, while PhyPrompt~\citep{wu2026phypromptrlbasedpromptrefinement} trains a physics-aware rewriter with supervised fine-tuning and GRPO. These works highlight the importance of prompt enhancement as an interface between user requests and video generators.

Modern commercial T2V systems also increasingly rely on proprietary prompt enhancers. For example, HappyHorse~1.1~\citep{happyhorse}, Kling~3.0~\citep{klingai2026}, MiniMax-H3~\citep{minimaxh3}, Seedance~2.0~\citep{seedance2026seedance20advancingvideo}, and Seedance~2.5~\citep{bytedance2026seedance25} employ native prompt-enhancement components whose model designs and training recipes remain largely undisclosed. MiniMax-H3 provides H3-Context-IR as a separately accessible prompt enhancer paired with H3-Base, whereas the prompt enhancers of other systems are integrated into their end-to-end generation APIs.

\section{Conclusion}
This work reframes prompt enhancement for modern video generators. As generators process longer contexts, follow complex instructions, and synthesize videos of up to 30 seconds, the prompt is no longer a decorated caption but the blueprint that orchestrates actions, camera choreography, audio, and narrative across shots and through time before pixel rendering. We introduced \method{} as a scalable realization of this cinematic planning architecture. It learns realized cinematic structure from real-world videos through video-grounded reverse SFT and faithfully preserves user requirements throughout the unfolding plan through SC-GRPO. Scaling \method{} from 4B to 397B consistently improves semantic fidelity and downstream video quality. Ablations show that reverse construction outperforms forward rewriting and that SC-GRPO improves semantic fidelity across model scales, while format-adapted \method{} outputs remain effective across video generators. These findings lay a reliable foundation for the next stage of prompt enhancement development.

\section{Acknowledgements}

We sincerely thank Junjie He, Xinhua Cheng, Zeyinzi Jiang, Xiaowen Li, Wei Wang$\sim$1, Tianyi Gui, Xiaoyi Bao, Wenting Shen, Tianxing Wang, Ang Wang, Weize Duan, Wei Wang$\sim$2, Zhi-Fan Wu, Chaojie Mao, and Lei Shang for their valuable support and contributions to this work.

\clearpage
\bibliographystyle{unsrtnat}
\bibliography{paper}

\clearpage
\clearpage
\appendix

\section{Video Data Collection and Curation}
\label{app:data_collection}

This section describes the real-world video corpus used to construct video-grounded supervision. Section~\ref{app:data_scope} introduces the collection scope. Section~\ref{app:data_organization} describes the organization of shot- and sequence-level clips. Section~\ref{app:data_filtering} presents the quality filtering and curation criteria. Section~\ref{app:data_corpus} summarizes the resulting corpus.

\subsection{Collection Scope}
\label{app:data_scope}

Cinematic planning requires supervision that captures both local realization and temporal organization. A detailed individual shot provides evidence about subject behavior, composition, camera work, lighting, and sound, whereas a sequence reveals how these elements develop through transitions, editing, and narrative progression. We therefore retain both shot-level clips and multi-shot sequences.

The source videos are publicly available or licensed for use. Long-form videos are divided into clips of no more than \(30\) seconds, allowing each sample to preserve detailed audiovisual content while remaining suitable for structured cinematic captioning.

\subsection{Shot- and Sequence-Level Organization}
\label{app:data_organization}

At the shot level, the corpus emphasizes semantically complete and visually expressive moments. The relevant signals include character actions and emotions, lighting and color, camera configuration and movement, music, speech, sound effects, animation-specific appearance and motion, and motion-graphics design. These clips provide dense evidence about how cinematic elements are combined within a single shot.

At the sequence level, the emphasis shifts to relationships among temporally ordered shots. The retained sequences expose changes in action, scene, camera, and audio, together with the transitions and editing patterns that connect them. They therefore provide direct examples of how individually meaningful shots are arranged into a coherent narrative progression.

This organization supplies complementary supervision: shot-level clips capture fine-grained cinematic realization, while sequence-level clips capture long-range structure.

\subsection{Quality Filtering and Curation}
\label{app:data_filtering}

We first filter the collected clips for technical validity, visual quality, and motion quality. The remaining samples are then curated according to their temporal structure.

For shot-level clips, curation considers the clarity and completeness of the central event together with the quality of character performance, visual presentation, camera behavior, audio-visual content, animation, and motion graphics.

For multi-shot sequences, curation considers whether the sequence is understandable as an independent unit, whether its ordered shots form a meaningful narrative progression, and whether it uses sufficiently rich cinematic organization. Content-specific properties are also considered where applicable, including visual-style consistency across animated shots and effective information delivery in motion-graphics sequences.

\subsection{Resulting Corpus}
\label{app:data_corpus}

The resulting corpus contains approximately \(1.05\mathrm{M}\) clips of up to \(30\) seconds across ten content dimensions. By combining fine-grained shot-level evidence with realized multi-shot organization, the corpus supports learning cinematic conditions at both local and long-range temporal scales. Appendix~\ref{app:reverse_construction} describes how these videos are converted into video-grounded cinematic targets.

\section{Video-Grounded Reverse Construction}
\label{app:reverse_construction}

This section describes how curated videos are converted into video-grounded cinematic targets. Section~\ref{app:caption_target} introduces the shared caption structure, Section~\ref{app:category_captioning} describes category-adaptive captioning, and Section~\ref{app:caption_validation} presents caption quality control.

\subsection{Hierarchical Caption Structure}
\label{app:caption_target}

Each cinematic target \(y_i\) contains a video-level summary and temporally ordered shot-level descriptions with timestamps. The video-level summary captures the overall content, setting, visual style, narrative perspective, pacing, and audio design. Each shot describes its composition, subjects, actions, lighting, camera movement, transitions, dialogue, music, and sound effects.

\subsection{Category-Adaptive Captioning}
\label{app:category_captioning}

To accommodate diverse video content, we use differentiated captioning schemes. General videos emphasize comprehensive scene coverage and detail completeness. Large-motion videos focus on displacement, speed, body-posture dynamics, and fast-paced audio. Music-related videos jointly describe performance, movement rhythm, music style, lyrics, and audio-visual synchronization. Dialect videos require accurate speech transcription grounded in visual context, while rich-text videos capture visible text together with its position, style, and layout.

For animation, captions emphasize visual style, animation-specific motion, and audio characteristics. Advertisement and motion-graphics videos focus on visual design, motion pacing, and brand presentation. Visual-effects videos describe effect types and their integration with the scene. Videos with complex camera movement or long takes require descriptions of movement type, direction, speed, and narrative function.

The captioner applies distinct system prompts for different schemes and uses preprocessing tags produced by vision-language or lightweight specialized models to reinforce the relevant descriptive dimensions.

\subsection{Caption Quality Control}
\label{app:caption_validation}

We assess generated captions for structural completeness, timestamp validity, and consistency with the source video. Captions that fail these checks are regenerated. The retained captions serve as the video-grounded targets used for supervised fine-tuning.
\section{\bench{} Design and Construction}
\label{app:benchmark_design}

\bench{} contains 249 curated requests spanning diverse content categories, target durations, aspect ratios, and levels of request granularity. Section~\ref{app:benchmark_distribution} presents the benchmark distribution, while Section~\ref{app:benchmark_examples} provides representative intent-level, scene-level, and shot-level requests.

\subsection{Benchmark Coverage and Distribution}
\label{app:benchmark_distribution}

\begin{figure}[h!]
    \centering
    \includegraphics[width=\linewidth]{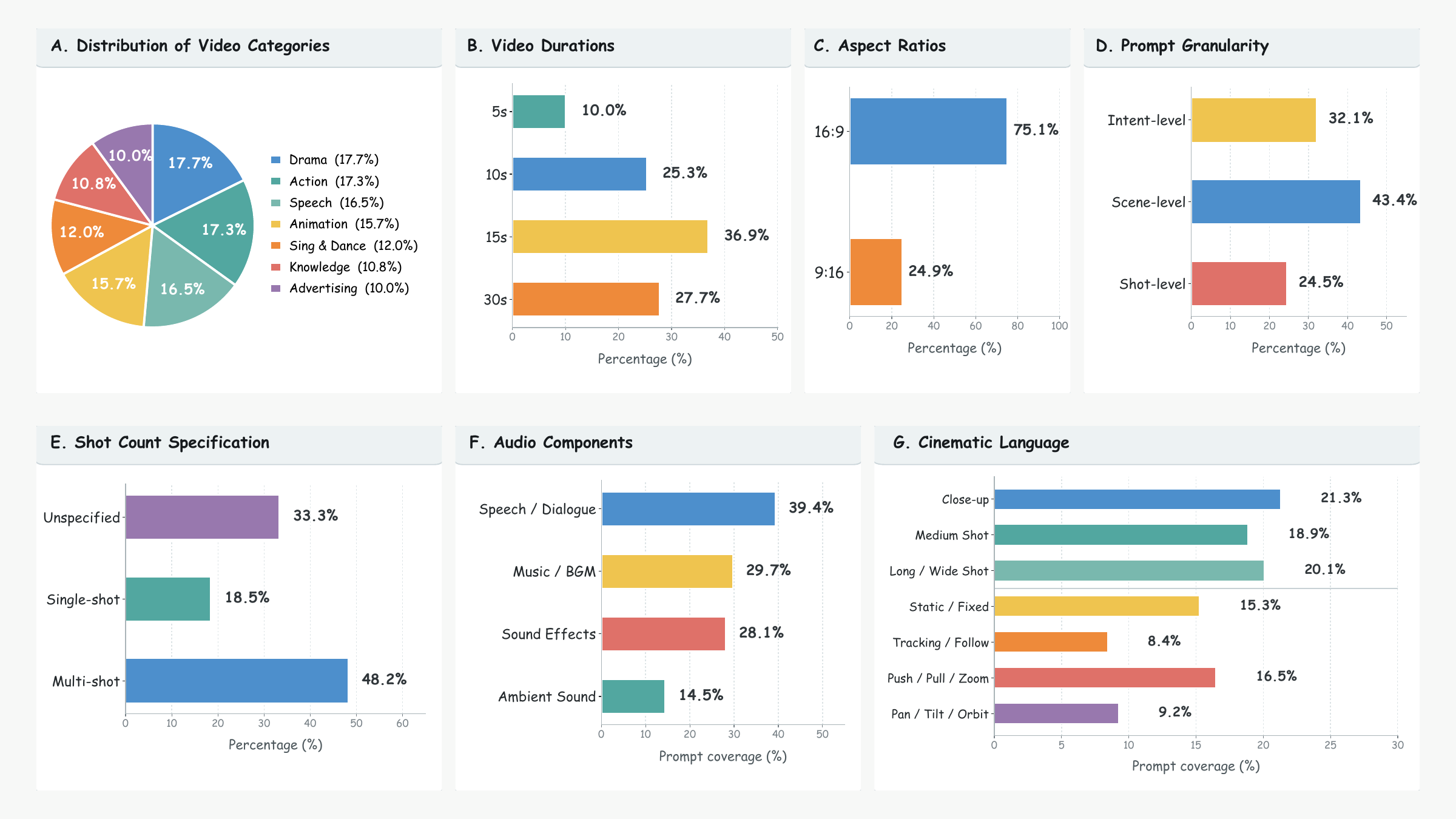}
    \caption{Distribution of the 249 requests in \bench{}.  The panels report
    video category, target duration, aspect ratio, prompt granularity, shot-count
    specification, audio components, and cinematic language.}
    \label{fig:benchmark_distribution}
\end{figure}

Figure~\ref{fig:benchmark_distribution} summarizes the distribution of the 249 requests in \bench{}. The benchmark covers seven content categories, four target durations from 5 to 30 seconds, two aspect ratios, and three request-granularity levels. It further characterizes shot-count specifications, audio components, and cinematic language across the requests. Together, these dimensions reflect practical text-to-video generation scenarios and requirements.

\subsection{Representative Requests Across Granularity Levels}
\label{app:benchmark_examples}

The following nine requests are taken from \bench{}, with three examples at each granularity level. They illustrate the diversity of the benchmark across categories, durations, and levels of request detail.

\noindent\begin{minipage}{\linewidth}
\textbf{Intent-level examples.}

\noindent\textit{Drama, 10 seconds, 16:9.}
\begin{quote}\small
Cinematic realistic style, a company corridor in the early morning, a mid-long
shot of an intern working overtime alone, constantly complaining, being
overheard by the boss.
\end{quote}
\end{minipage}

\noindent\textit{Speech, 5 seconds, 16:9.}
\begin{quote}\small
Generate a historical court drama dialogue video with 3 shots.  The content is
about an imperial concubine aggressively scolding a maid, and the maid
responding aggrievedly.
\end{quote}

\noindent\textit{Animation, 15 seconds, 9:16.}
\begin{quote}\small
3D exquisite realistic animation, in front of a Gothic church under the
moonlight, a little girl in a Gothic nun's attire holds a thorny rose and makes
eye contact with a plaster statue of Jesus, cool color tone, exquisite and
beautiful.
\end{quote}

\noindent\textbf{Scene-level examples.}

\noindent\textit{Drama, 10 seconds, 16:9.}
\begin{quote}\small
A young man with the power to sense and purify evil energies walks alone
through a deserted alley on a rainy night.  Suddenly, he is entangled by a
ferocious evil spirit lurking in the dark.  He faces the roaring, pouncing evil
spirit alone, fighting against the malevolent forces by himself to protect the
peace of the human world.
\end{quote}

\noindent\textit{Action, 30 seconds, 16:9.}
\begin{quote}\small
A wandering monk wearing a bamboo hat is traveling when he witnesses a young
woman being harassed by thugs.  He immediately rushes forward, uses the bamboo
flute in his hand to fight them off and rescues the woman.  After accepting her
thanks, he turns and walks away, playing the bamboo flute as he goes.
\end{quote}

\noindent\textit{Animation, 15 seconds, 16:9.}
\begin{quote}\small
Classic fairytale princess style 3D animation.  In a glass vase in a flower
shop, a small white flower quietly blooms, and a glowing flower fairy appears
from within the petals.  She slowly floats among the flower stands, looking
around at the room full of fresh flowers and water droplets.  Suddenly, she
hears the shop's doorbell ring, immediately returns inside the flower bud, and
the flower bud closes again.
\end{quote}

\noindent\textbf{Shot-level examples.}

\noindent\textit{Drama, 10 seconds, 16:9.}
\begin{quote}\small
Shot 1: Long shot, modern realistic style, a playground.  Afternoon sun casts
dappled light and shadows through tree leaves, with students walking and
jogging in the background.  The camera slowly pushes in, focus gradually shifts
from the light spots on the playground floor to the running students in the
distance.  The foreground leaves are slightly blurred to create a sense of
spatial depth.

Shot 2: Mid shot, tracking.  The male protagonist, carrying a backpack, walks
towards the school building.  The camera follows from his side-rear, with a
stabilizer for smooth tracking.  A slight high angle captures the rhythm of his
steps and his elongated shadow.  The background is blurred to make the subject
stand out.

Shot 3: Mid shot pushing in to a close-up.  The female protagonist is sitting
on a bench, reading a book.  Dappled sunlight falls on the tips of her hair and
the book pages.  The camera slowly pushes in from a mid shot to a side-profile
close-up.  A slight handheld shake is added for a more lifelike feel.  A shallow
depth of field focuses on the female protagonist's expression, with the
background softly blurred.

Shot 4: Close shot.  The male protagonist walks up to the female protagonist
and hands her a bottle of water.  The camera pans horizontally from the male
protagonist's hand to the female protagonist's hand, keeping both centered in
the frame.  A slight focus pull transitions between them, capturing the moment
their hands touch.  The background remains slightly blurred.
\end{quote}

\noindent\begin{minipage}{\linewidth}
\textit{Speech, 5 seconds, 16:9.}
\begin{quote}\small
A dialogue scene of a long-awaited reunion in an old mansion from the Republic
of China era.  The text and tone must exactly match the instructions.
Shot 1: A two-person mid-shot, static camera.  A girl in a simple and elegant
cheongsam (in the calm tone of a young lady from a wealthy family): ``You still
know to come back.'' A man in a dark Zhongshan suit (in the sincere tone of a
mature and steady young man): ``I haven't forgotten our promise.''
Shot 2: Cut to a close-up of the girl's side profile.  The girl (emotionally
overwhelmed, crying, voice trembling and choked): ``All these years, I've been
waiting for you.''
\end{quote}
\end{minipage}

\noindent\textit{Speech, 30 seconds, 16:9.}
\begin{quote}\small
On a late autumn evening, in a quiet street alley, a tall ginkgo tree gently
sheds its golden leaves.  Pale, warm evening light filters through the  
branches, casting mottled bright spots.  The ambient sound is of a low wind and
the subtle rustling of leaves scraping the ground.  The camera uses a fixed
medium shot, where a woman in a simple, light-colored knitted sweater with her
long hair hanging naturally stands quietly by the ginkgo tree, her gaze fixed
on the end of the alley.  Then, the frame slowly cuts to a close-up of her
fingers gently twirling a leaf, capturing the small and restrained action.
Next, it slowly transitions to a wide shot; the woman turns slightly to the
side and walks with a composed pace deeper into the street alley, her silhouette
gradually receding.  The camera slowly pulls back, finally stopping on a quiet
picture of her back intertwined with the long alley in late autumn.  Narrator
(in a calm tone): ``The wind brushes the long street, yellow leaves fall
silently.  Those fragments buried deep in the years, when recalled, leave only
a touch of warmth and a bit of melancholy that was never spoken.''
\end{quote}

These examples illustrate how request detail varies across the three granularity levels. Intent-level requests primarily specify a high-level scenario with limited temporal detail. Scene-level requests describe the progression of events without detailed shot-by-shot instructions. Shot-level requests explicitly organize content into shots and specify elements such as shot scale, camera movement, focus, action, and dialogue.

\section{Additional Quantitative Results}
\label{app:additional_quantitative_results}

This section provides additional quantitative results that supplement Table~\ref{tab:short_medium_result} and Figure~\ref{fig:\bench{}_breakdown}.  We report baseline-specific preference results and exact breakdowns by duration and request granularity on the \(5\)-\(15\)-second subset of \bench{}.

\subsection{Baseline-Specific Preference Results}
\label{app:complete_pairwise_results}

Table~\ref{tab:direct_pairwise_duration} reports preference results relative to individual external baselines. Each block compares an external system with the original-request baseline and \method{}-397B, with the gray row indicating the downstream generator used by the latter two. Scores aggregate valid pairwise judgments within each block.

\begin{center}
\begin{minipage}{\textwidth}
\refstepcounter{table}\label{tab:direct_pairwise_duration}
\centering
Table~\thetable: Baseline-specific preference scores across seven content categories and overall.\par
\vspace{2pt}
\scriptsize
\setlength{\tabcolsep}{1.4pt}
\renewcommand{\arraystretch}{1.00}
\begin{tabular*}{\textwidth}{@{\extracolsep{\fill}}l|cccccccc@{}}
\toprule
Method & Action & Anim. & Speech & Ad. & Sing \& Dance & Drama & Know. & Overall \\
\midrule
\rowcolor{blue!10}
\multicolumn{9}{c}{\textit{LTX-2.5 vs.\ Ours}} \\
LTX-2.5 & 6.78 & 10.83 & \underline{36.21} & 19.23 & 5.56 & 16.10 & \underline{26.32} & 17.34 \\
\rowcolor{gray!12}
\multicolumn{9}{c}{\textit{Downstream generator: Wan3.0's video generator}} \\
Original Request & \underline{43.33} & \underline{48.33} & 35.34 & \underline{38.46} & \underline{47.37} & \underline{35.00} & 11.54 & \underline{37.85} \\
+\method{}-397B & \textbf{50.85} & \textbf{57.50} & \textbf{40.52} & \textbf{48.68} & \textbf{52.63} & \textbf{51.69} & \textbf{50.00} & \textbf{50.28} \\
\midrule
\rowcolor{blue!10}
\multicolumn{9}{c}{\textit{Kling 3.0 vs.\ Ours}} \\
Kling 3.0 & 9.32 & 10.00 & \underline{35.34} & 16.67 & 26.25 & 18.97 & \underline{17.57} & 18.95 \\
\rowcolor{gray!12}
\multicolumn{9}{c}{\textit{Downstream generator: Wan3.0's video generator}} \\
Original Request & \underline{40.00} & \underline{42.50} & 26.32 & \underline{32.05} & \underline{35.00} & \underline{32.50} & 12.82 & \underline{32.54} \\
+\method{}-397B & \textbf{50.00} & \textbf{62.50} & \textbf{44.92} & \textbf{57.89} & \textbf{48.75} & \textbf{50.86} & \textbf{56.58} & \textbf{52.84} \\
\midrule
\rowcolor{blue!10}
\multicolumn{9}{c}{\textit{HappyHorse 1.1 vs.\ Ours}} \\
HappyHorse 1.1 & 12.07 & 17.50 & 26.27 & \underline{31.58} & 8.75 & \underline{24.56} & \underline{35.53} & 21.71 \\
\rowcolor{gray!12}
\multicolumn{9}{c}{\textit{Downstream generator: Wan3.0's video generator}} \\
Original Request & \underline{36.44} & \underline{40.00} & \underline{34.75} & 24.36 & \underline{47.50} & 24.14 & 3.85 & \underline{31.07} \\
+\method{}-397B & \textbf{44.92} & \textbf{60.83} & \textbf{50.00} & \textbf{50.00} & \textbf{51.25} & \textbf{52.54} & \textbf{46.15} & \textbf{51.14} \\
\midrule
\rowcolor{blue!10}
\multicolumn{9}{c}{\textit{MiniMax-H3 vs.\ Ours}} \\
MiniMax-H3 & 11.54 & 25.00 & 22.50 & \underline{33.33} & 25.00 & \underline{39.17} & \underline{42.11} & 27.83 \\
\rowcolor{gray!12}
\multicolumn{9}{c}{\textit{Downstream generator: Wan3.0's video generator}} \\
Original Request & \underline{35.96} & \underline{37.07} & \underline{44.07} & 28.95 & \underline{37.50} & 30.00 & 12.82 & \underline{33.33} \\
+\method{}-397B & \textbf{42.73} & \textbf{59.48} & \textbf{53.39} & \textbf{46.15} & \textbf{45.00} & \textbf{45.83} & \textbf{48.72} & \textbf{49.14} \\
\midrule
\rowcolor{blue!10}
\multicolumn{9}{c}{\textit{Seedance 2.0 vs.\ Ours}} \\
Seedance 2.0 & \textbf{41.11} & \underline{44.90} & 35.96 & \underline{33.78} & 29.03 & \underline{43.75} & \underline{30.56} & \underline{37.94} \\
\rowcolor{gray!12}
\multicolumn{9}{c}{\textit{Downstream generator: Wan3.0's video generator}} \\
Original Request & 27.36 & 22.22 & \underline{38.60} & 31.58 & \underline{31.43} & 22.41 & 5.26 & 25.98 \\
+\method{}-397B & \underline{38.46} & \textbf{52.73} & \textbf{42.24} & \textbf{47.30} & \textbf{38.89} & \textbf{50.86} & \textbf{39.47} & \textbf{44.76} \\
\bottomrule
\end{tabular*}
\end{minipage}
\end{center}

\subsection{Fine-Grained Preference Breakdowns}
\label{app:fine_grained_preference_results}

Tables~\ref{tab:preference_duration_intervals}
and~\ref{tab:preference_granularity_intervals} list the exact values represented
by the markers and intervals in the left panel of
Figure~\ref{fig:\bench{}_breakdown}.  Each entry follows the order
\(S_{\mathrm{lower}} / S / S_{\mathrm{upper}}\).  \method{}-397B attains the
highest central score at every duration and request granularity.

\begin{center}
\begin{minipage}{\textwidth}
\refstepcounter{table}\label{tab:preference_duration_intervals}
\centering
Table~\thetable: Exact preference intervals by duration.  Entries are
\(S_{\mathrm{lower}} / S / S_{\mathrm{upper}}\).\par
\vspace{2pt}
\scriptsize
\setlength{\tabcolsep}{3.0pt}
\renewcommand{\arraystretch}{1.10}
\begin{tabular*}{\textwidth}{@{\extracolsep{\fill}}lccc@{}}
\toprule
Method & 5 seconds & 10 seconds & 15 seconds \\
\midrule
LTX-2.5 & 17.29 / 18.69 / 20.09 & 16.04 / 18.47 / 20.90 & 14.39 / 16.10 / 17.80 \\
Kling 3.0 & 26.27 / 31.11 / 35.94 & 19.68 / 23.47 / 27.26 & 14.07 / 16.10 / 18.12 \\
HappyHorse 1.1 & 15.67 / 20.28 / 24.88 & 20.65 / 24.00 / 27.36 & 23.20 / 26.80 / 30.41 \\
MiniMax-H3 & 25.00 / 29.09 / 33.17 & 31.77 / 36.46 / 41.16 & 33.93 / 38.30 / 42.67 \\
Seedance 2.0 & 42.11 / 46.20 / 50.29 & 36.23 / 41.40 / 46.58 & 39.94 / 44.31 / 48.69 \\
\midrule
\rowcolor{gray!12}
\multicolumn{4}{c}{\textit{Downstream generator: Wan3.0's video generator}} \\
Original request & 39.63 / 46.08 / 52.53 & 27.70 / 33.63 / 39.57 & 26.09 / 30.59 / 35.08 \\
+\method{}-4B & 41.74 / 48.17 / 54.59 & 35.83 / 41.27 / 46.70 & 38.61 / 44.00 / 49.38 \\
+\method{}-9B & 43.84 / 50.91 / 57.99 & 39.32 / 44.40 / 49.47 & 41.12 / 45.78 / 50.43 \\
+\method{}-35B & 42.59 / 50.69 / 58.80 & 41.84 / 48.76 / 55.67 & 40.97 / 46.45 / 51.93 \\
+\method{}-397B & 48.37 / 56.74 / 65.12 & 42.37 / 49.91 / 57.45 & 43.18 / 49.43 / 55.68 \\
\bottomrule
\end{tabular*}
\end{minipage}
\end{center}

\begin{center}
\begin{minipage}{\textwidth}
\refstepcounter{table}\label{tab:preference_granularity_intervals}
\centering
Table~\thetable: Exact preference intervals by request granularity.  Entries
are \(S_{\mathrm{lower}} / S / S_{\mathrm{upper}}\).\par
\vspace{2pt}
\scriptsize
\setlength{\tabcolsep}{3.0pt}
\renewcommand{\arraystretch}{1.10}
\begin{tabular*}{\textwidth}{@{\extracolsep{\fill}}lccc@{}}
\toprule
Method & Intent-level & Scene-level & Shot-level \\
\midrule
LTX-2.5 & 17.23 / 19.21 / 21.19 & 12.92 / 14.39 / 15.87 & 15.70 / 18.18 / 20.66 \\
Kling 3.0 & 17.35 / 20.66 / 23.98 & 16.45 / 18.79 / 21.12 & 20.44 / 24.03 / 27.62 \\
HappyHorse 1.1 & 21.18 / 25.42 / 29.65 & 21.78 / 25.00 / 28.22 & 20.51 / 23.74 / 26.97 \\
MiniMax-H3 & 33.28 / 38.37 / 43.45 & 31.05 / 35.05 / 39.05 & 30.87 / 34.84 / 38.80 \\
Seedance 2.0 & 41.61 / 46.83 / 52.05 & 37.89 / 42.21 / 46.53 & 35.10 / 39.23 / 43.36 \\
\midrule
\rowcolor{gray!12}
\multicolumn{4}{c}{\textit{Downstream generator: Wan3.0's video generator}} \\
Original request & 29.07 / 33.89 / 38.70 & 27.86 / 32.29 / 36.72 & 28.53 / 35.87 / 43.21 \\
+\method{}-4B & 39.97 / 45.96 / 51.95 & 37.84 / 42.05 / 46.25 & 34.95 / 41.67 / 48.39 \\
+\method{}-9B & 42.90 / 48.28 / 53.66 & 40.29 / 44.87 / 49.45 & 38.01 / 43.53 / 49.06 \\
+\method{}-35B & 41.94 / 49.33 / 56.72 & 42.83 / 48.35 / 53.86 & 38.75 / 44.44 / 50.14 \\
+\method{}-397B & 44.82 / 52.44 / 60.06 & 43.91 / 49.26 / 54.61 & 40.98 / 49.32 / 57.65 \\
\bottomrule
\end{tabular*}
\end{minipage}
\end{center}

\end{document}